\documentclass{article}
\usepackage{spconf,amsmath,graphicx}
\usepackage{multirow}
\usepackage{booktabs,subcaption}
\usepackage{amsmath,amsfonts,dsfont,cleveref}

\usepackage{bm,xspace}
\usepackage{color}
\usepackage{CJKutf8}

\newcommand{\ctext}[1]{\begin{CJK}{UTF8}{gbsn}\small{#1}\end{CJK}}

\newcommand{\tabincell}[2]{\begin{tabular}{@{}#1@{}}#2\end{tabular}}

\newcommand{\system}{{\tt AdvGraph}\xspace}

\crefname{equation}{Eq.}{Eqs.}
\crefname{table}{Table.}{Tables.}
\crefname{figure}{Fig.}{Figs.}
\crefname{algocf}{alg.}{algs.}

\title{ENHANCING MODEL ROBUSTNESS BY INCORPORATING ADVERSARIAL KNOWLEDGE INTO SEMANTIC REPRESENTATION}

\name{Jinfeng Li$^{1}$\sthanks{Contacted e-mail: jinfengli.ljf@alibaba-inc.com} \quad Tianyu Du$^{2}$ \quad Xiangyu Liu$^{1}$ \quad Rong Zhang$^{1}$ \quad Hui Xue$^{1}$ \quad Shouling Ji$^{2}$}

\address{$^{1}$Alibaba Group, Hangzhou, China \\$^{2}$  Zhejiang University, Hangzhou, China}

\begin{document}
\ninept
\maketitle
\begin{abstract}
Despite that deep neural networks (DNNs) have achieved enormous success in many domains like natural language processing (NLP), they have also been proven to be vulnerable to maliciously generated adversarial examples. 
Such inherent vulnerability has threatened various real-world deployed DNNs-based applications.
To strength the model robustness, several countermeasures have been proposed in the English NLP domain and obtained satisfactory performance.
However, due to the unique language properties of Chinese, it is not trivial to extend existing defenses to the Chinese domain.
Therefore, we propose \system, a novel defense which enhances the robustness of Chinese-based NLP models by incorporating adversarial knowledge into the semantic representation of the input.
Extensive experiments on two real-world tasks show that \system exhibits better performance compared with previous work: 
(i) effective -- it significantly strengthens the model robustness even under the adaptive attacks setting without negative impact on model performance over legitimate input;
(ii) generic -- its key component, i.e., the representation of connotative adversarial knowledge is task-agnostic, which can be reused in any Chinese-based NLP models without retraining;
and (iii) efficient -- it is a light-weight defense with sub-linear computational complexity, which can guarantee the efficiency required in practical scenarios.
\end{abstract}

\vspace{-0.15cm}
\section{INTRODUCTION}
\vspace{-0.15cm}
\label{sec:intro}
Deep neural networks (DNNs) have recently revolutionized natural language processing (NLP) with their impressive performance on many tasks, including text classification \cite{kim2014convolutional,devlin2018bert,zhou2016attention}, machine translation \cite{bahdanau2014neural,cho2014properties} and question answering \cite{bonadiman2017effective,unlu2019question}. Such advances in DNNs have led to broad deployment of systems on important tasks in physical world.
However, recent studies have revealed that DNNs are inherently vulnerable to adversarial examples that are maliciously crafted by adding small perturbations into benign input that can trigger the target DNNs to misbehave \cite{szegedy2013intriguing,li2019textbugger,behjati2019universal}.
Such vulnerability has raised great concerns about the security as well as the deployment of DNNs in real-world tasks especially the security-sensitive tasks. 

\begin{figure}[t]
\setlength{\abovecaptionskip}{2pt}
\centering
\includegraphics[width=0.455\textwidth]{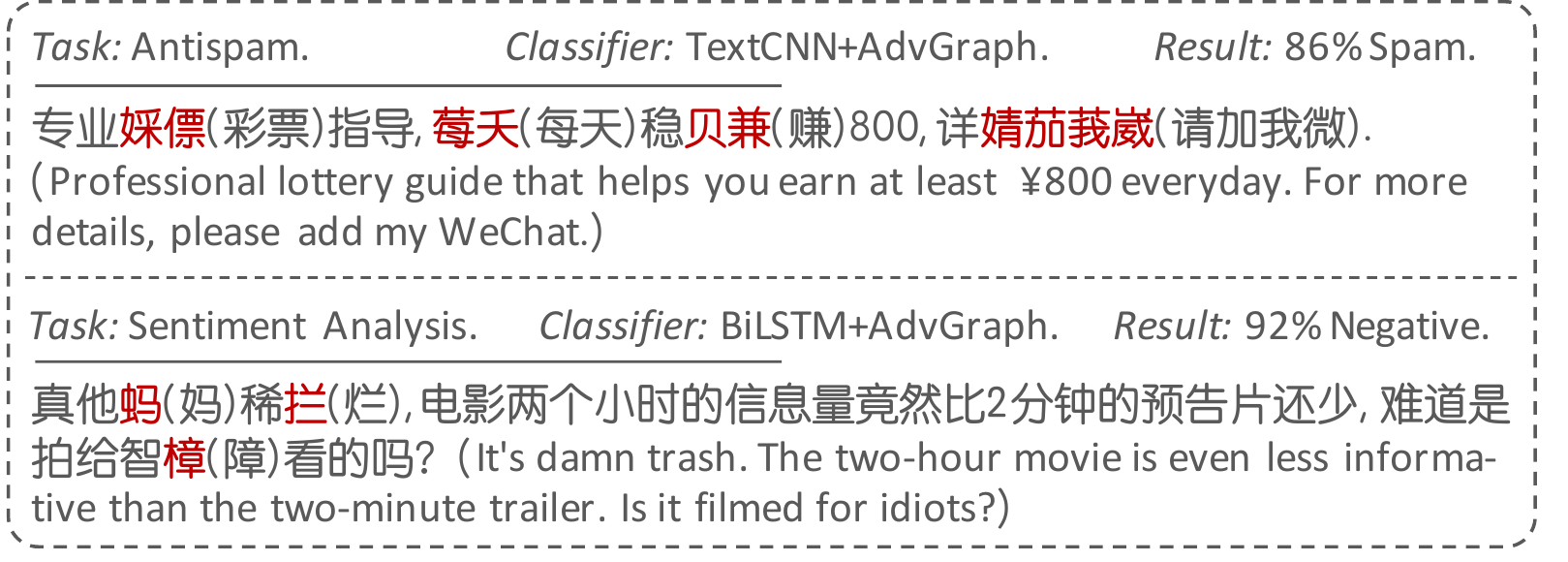}
\caption{User-generated obfuscated texts which bypass the commercial text moderation systems of Baidu and Netease. The characters in brackets are the original ones while those in red are the variants.}
\label{fig:real_example}
\vspace{-0.6cm}
\end{figure}

In the meantime, DNNs-based text classification has been the backbone technique behind online toxic content censorship systems \cite{karan2018cross,nobata2016abusive}, which have been widely used to automatically detect toxic user-generated content for purifying the online social network and have achieved great success in replacing the time-consuming and laborious manual censorship.
Nevertheless, it cannot be ignored that there are many malicious netizens in online social networks who usually obfuscate their insulting comments by replacing the toxic words with their variants (also known as \textit{morphs}) to evade the detection of censorship systems \cite{li2020textshield}, and the situation is even worse on Chinese social media \cite{chen2013tweeting}. 
For instance, to make their insulting comments evasive, malicious netizens may obfuscate some toxic words in their comments with the corresponding variants, such as substituting ``\ctext{智障}'' (idiot) with ``\ctext{智樟}'' as shown in \cref{fig:real_example}.
These variants are usually visually or phonetically similar to their original words, which can retain the toxic meaning from the human perspective due to the powerful cognitive and perceptual capabilities of human beings.
However, due to the inherent vulnerability of DNNs to adversarial examples, the DNNs-based censorship systems can be easily deceived into making wrong decisions on these adversarial texts.
It was reported that major social media like Twitter, Facebook, and Sina Weibo were all criticized for not doing enough to curb the diffusion of such toxic content and under pressure to cleanse their platforms \cite{facebook_criticized,weibo_criticized}.
Therefore, it is urgent to develop the corresponding defense countermeasures for improving the robustness of real-world deployed systems against the aforementioned adversarial texts.

To this end, several defense methods have been proposed in the English NLP domain, which can be summarized into \textit{adversarial training} and \textit{spelling correction}.
Specifically, Wang \textit{et al.} \cite{wang2018robust} and Cheng \textit{et al.} \cite{cheng2020advaug} proposed to retrain the NLP model with diversified adversarial training data and showed a marginal increase in robustness. 
Zhou \textit{et al.} \cite{zhou2019learning} proposed a spelling correction-based framework for blocking adversarial texts, which first identifies the perturbed tokens by a discriminator and then recovers the tokens from discriminated perturbations by a masked language model objective with contextualized language modeling.
Similarly, Li \textit{et al.} \cite{li2019textbugger} adopted the context-aware spelling correction method to mitigate editorial adversarial attacks and achieved satisfactory performance.

However, although these methods have been shown to be effective in enhancing the robustness of English-based NLP models, it is still intractable to extend them to the Chinese NLP domain due to the following unique properties of Chinese:
(i) \textit{complexity} -- Chinese is a logographic language without word delimiters, in which each character is individually meaningful and the meaning of each character changes dramatically when the context changes, which makes Chinese NLP models inherently more vulnerable;
(ii) \textit{sparsity} and \textit{diversity} -- there is an extremely large and sparse character space (i.e., there are more than 50,000 Chinese characters, of which only about 3,500 are commonly used \footnote{https://en.wikipedia.org/wiki/Chinese\_characters}), and each character might be perturbed by various variation strategies (e.g., glyph- and phonetic-based strategies) \cite{li2020textshield}, which makes the variants more diverse and sparse and thus limits the efficacy of common defenses;
and (iii) \textit{dynamicity} -- in the context of toxic content detection on Chinese online social media or e-commercial platforms, the arms race between adversaries and defenders is extremely fierce \cite{li2019spam}, and static defense methods like adversarial training and spelling correction are usually only effective in handling known variants and are still vulnerable to new attacks.
Hence, the defense against adversarial texts in the Chinese NLP domain remains a challenging and unsolved problem.

\begin{figure}[t]
\setlength{\abovecaptionskip}{2pt}
\centering
\includegraphics[width=8.0cm]{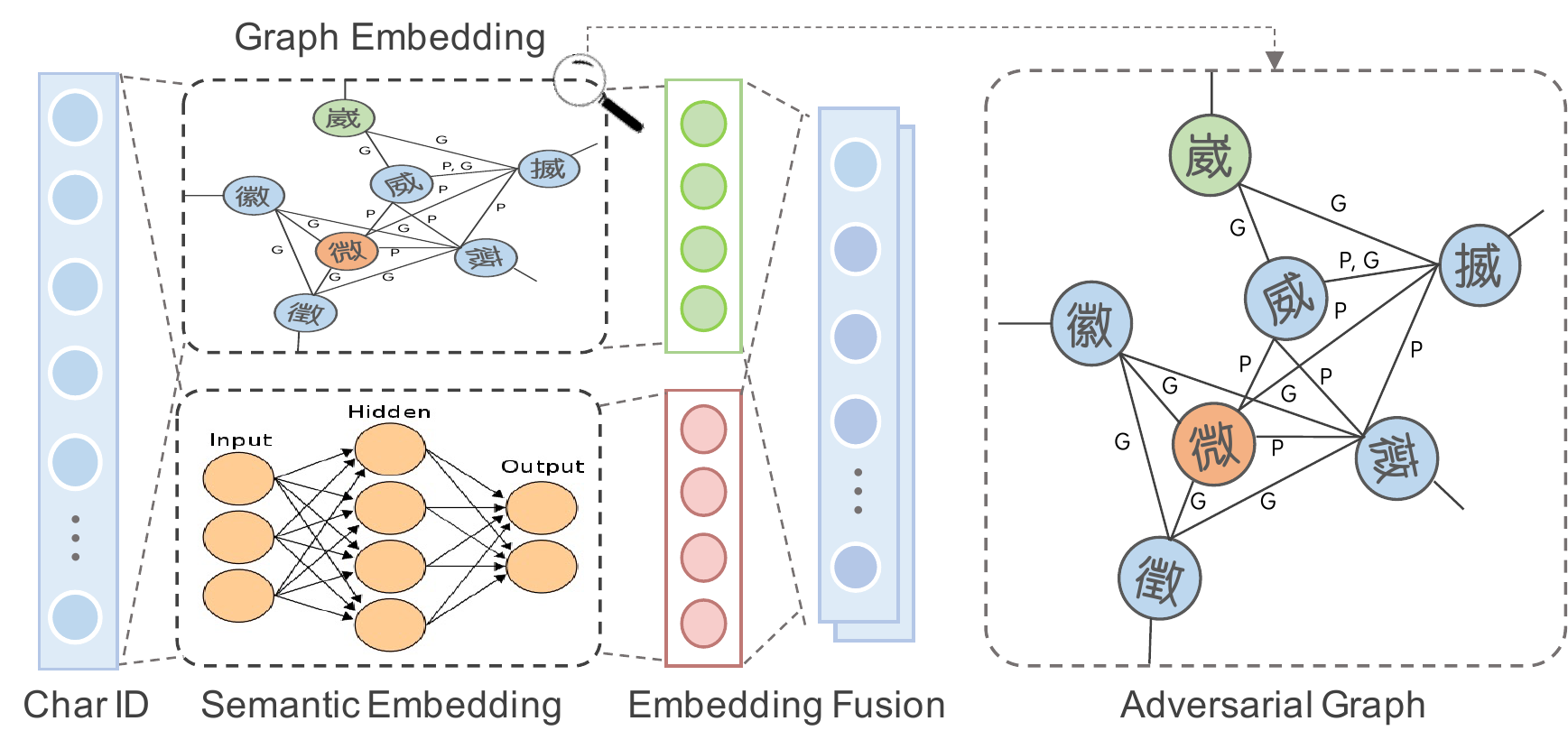}
\caption{The framework of our defense approach. The letters ``P'' and ``G'' in adversarial graph denote the phonetic-based and glyph-based variation relationship, respectively.}
\label{fig:framework}
\vspace{-0.55cm}
\end{figure}

To tackle the aforementioned challenges, in this paper, we propose \system, a novel defense to enhance the inherent robustness of Chinese-based NLP models by incorporating adversarial knowledge into the semantic representation of input texts (shown in \Cref{fig:framework}). 
At a high level, we first construct an undirected adversarial graph based on the glyph and phonetic similarity of Chinese characters to model the adversarial relationship between characters explicitly.
The intuition behind is that real-world adversarial texts are usually generated utilizing the glyph-based or phonetic-based perturbation \cite{liu2011visually}.
Then, we leverage the graph embedding scheme to learn the node representation of the adversarial graph for capturing the prior adversarial knowledge between characters. Finally, we incorporate the node representation into the semantic representation through multimodal fusion, and the fused semantic-rich representation is then ready for the downstream tasks.
Through extensive experiments, we show that \system can greatly improve the inherent model robustness even under the adaptive attack setting without negative impact on the model performance over benign texts, which outperforms the baselines by a significant margin.

\vspace{-0.25cm}
\section{PROPOSED METHOD}
\label{sec:method}

\vspace{-0.15cm}
\subsection{Problem definition}
\vspace{-0.15cm}
Consider a Chinese-based text classifier $\mathcal{F}\!:\mathcal{X} \!\to\! \mathcal{Y}$ which maps the input from the feature space $\mathcal{X}$ to the label space $\mathcal{Y}$, an adversary who has query access to the classification confidence returned by $\mathcal{F}$, aims to generate an adversarial text $\bm{x_{adv}}$ from a legitimate input $\bm{x}\in\mathcal{X}$ that contains $N$ characters (i.e., $\bm{x}\!=\!\{x_1, x_2, \cdots, x_N\}$) with the ground-truth label $y\in\mathcal{Y}$, so that $\mathcal{F}(\bm{x_{adv}})\neq\mathcal{F}(\bm{x})$.
In this paper, we aim to defend against such attacks by incorporating the adversarial knowledge learnt from an adversarial graph into the semantic representation of input. Then, our defense is formalized as 
\begin{equation}
\setlength{\abovedisplayskip}{4pt}
\setlength{\belowdisplayskip}{4pt}
    \mathcal{F}(\bm{x_{adv}}) = \mathcal{F}_f(E_{s}(\bm{x_{adv}})\oplus E_g(\bm{x_{adv}})) = \mathcal{F}(\bm{x}) = y,
\end{equation}
where $\mathcal{F}_f(\cdot)$ is the classification function of $\mathcal{F}$, $E_{s}(\bm{x_{adv}})$ is the semantic representation of $\bm{x_{adv}}$, $E_g(\bm{x_{adv}})$ is the adversarial representation learnt from an adversarial graph $G_{adv}$ and $\oplus$ is the fusing operation which incorporates the adversarial knowledge into the semantic representation to form a semantic-rich representation. 

\vspace{-0.15cm}
\subsection{Proposed defense}
\vspace{-0.1cm}
The framework of \system is presented in \Cref{fig:framework}, which is built upon adversarial graph, semantic embedding and multimodal fusion. 
Below, we will elaborate on each of the backbone techniques.

\textbf{Adversarial graph.}
\label{ssec:adv_graph}
The variation association among Chinese characters is usually a many-to-many relationship, i.e., a Chinese character may have multiple variants and it may also be a variant of different characters.
The relationship is symmetric since the two characters that are visually or phonetically similar are variants of each other.
Hence, we first leverage an undirected graph to model such adversarial relationships explicitly, in which each node denotes a Chinese character and an edge is formed if two characters have a variation relationship. 
Then, we learn the representation of the adversarial relationships among characters utilizing graph embedding.

\begin{figure}[t]
\setlength{\abovecaptionskip}{2pt}
\centering
\includegraphics[width=0.48\textwidth]{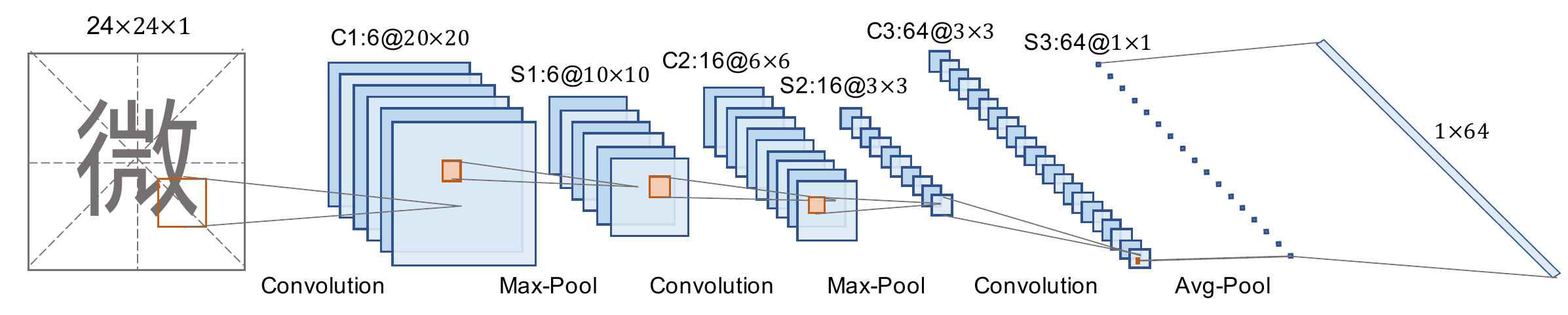}
\caption{Architecture of the glyph representation model.}
\label{fig:glyph_embedd_model}
\vspace{-0.55cm}
\end{figure}

\textbf{(1) Graph construction.} The adversarial graph $G_{adv}$ is intuitively built based on the glyph and phonetic similarity of Chinese characters since real-world adversarial text are usually generated based on the glyph- and phonetic-based perturbations \cite{liu2011visually}.
For the phonetic-based variation relation, we first convert each character into its \textit{pinyin} \footnote{\scriptsize Pinyin is the official Mandarin romanization system for standard Chinese.} form, and then calculate the edit distance between their pinyin. 
The characters are viewed as similar if the corresponding edit distance equals to 0 or 1 (only removal operation is considered), and an edge is then formed between them.
Different from phonetic similarity, the glyph similarity cannot be directly quantified since it is reflected through the visual perception of humans. 
It is also intractable to manually collect those characters that are visually similar due to the extremely large character space of Chinese. 
To tackle this challenge, we first convert each character into an image of size $24\!\times\!24$, then we implement a convolutional neural network g-CNN (shown in \cref{fig:glyph_embedd_model}) as done in \cite{li2020textshield} to learn the glyph representation of each character over the converted image. 
Different from \cite{li2020textshield}, our g-CNN is trained on 10,000 manually labelled triplets $(x_i, x^{+}_i, x^{-}_i)$ over 3,000 commonly used characters, in which $(x_i, x^{+}_i)$ and $(x_i, x^{-}_i)$ are the similar and dissimilar character pairs, respectively.
The glyph representation is learned by minimizing the triplet loss \cite{schroff2015facenet} defined as
\begin{equation}
\setlength{\abovedisplayskip}{4pt}
\setlength{\belowdisplayskip}{4pt}
    \mathcal{L} = \sum_{i}^{M=10,000} [\|h(x_i)-h(x^{+}_i)\|^{2}_{2} - \|h(x_i)-h(x^{-}_{i})\|^{2}_{2} + \alpha]_+,
\end{equation}
where $h(x_i)$ is the hidden representation of character $x_i$ and $\alpha$ is the hyper-parameter.
Then, the glyph-based variation relation is built by calculating the similarity between glyph representations $h(x_i)$, i.e, each character in $G_{adv}$ is connected to its top-10 visually similar characters.
The constructed $G_{adv}$ is illustrated in \cref{fig:framework}.

\textbf{(2) Graph embedding.}
To easily incorporate the complex adversarial knowledge into the semantic representation of the input, we first map the adversarial relationships in $G_{adv}$ to the feature space by learning the node representation using graph embedding scheme.
Concretely, we adopt the node2vec \cite{grover2016node2vec} to learn the node representation of $G_{adv}$.
Given $G_{adv}\!=\!(V,E)$, the mapping function $f\!:\!V\!\to\!\mathbb{R}^{d}$ from nodes set $V$ to feature space is learnt based on the idea of Skip-gram \cite{mikolov2013distributed} by maximizing the objective function
\begin{equation}
\setlength{\abovedisplayskip}{4pt}
\setlength{\belowdisplayskip}{4pt}
    \mathcal{L}(f, \theta) 
    =\sum_{x_i\in V}{log(\prod_{x_j\in N_S(x_i)}{p(x_j|f(x_i))})},
\end{equation}
where $\theta$ is the parameter of $f$, $p(x_j)|f(x_i))$ is the probability of observing the neighbor node $x_j$ for node $x_i$ conditioned on its feature representation, and $N_S(x_i)$ is the neighborhood of $x_i$ obtained by sampling strategies include both breath-first sampling (BFS) and depth-first sampling (DFS).
Particularly, BFS helps to model the potential adversarial relationships based on structural equivalence, i.e., nodes with similar structural roles in $G_{adv}$ are embedded closely since they might have the same variation situation in adversarial scenario.
DFS helps to model the direct adversarial relationships based on homophily, i.e., nodes that are highly interconnected or belong to similar clusters are represented closely since they usually express the same meaning in adversarial texts. 
Through the aforementioned framework, we can not only learn the adversarial knowledge in direct variation relationships, but also the knowledge in more complex variation relationships (e.g., secondary variation ``\ctext{微}''$\to$``\ctext{崴}'' shown in \cref{fig:real_example,fig:framework}, which ``\ctext{崴}'' does not have a direct relationship with ``\ctext{微}'' but is a common variant of ``\ctext{微}'' in spam advertisement), which is usually could not be handled by previous defense.

Note that the process for building $G_{adv}$ and learning node representation is task-agnostic and is done offline. Hence, the adversarial representation can be reused in any Chinese NLP models once learnt, which greatly improves the generality and efficiency of the defense.

\textbf{Semantic embedding.}
We can either use the classical Word2Vec models like CBoW and Skip-gram \cite{mikolov2013efficient} or directly utilize the pre-trained embedding of language models like BERT \cite{devlin2018bert} to obtain the semantic embedding.
To make a fair comparison for better verifying the efficacy of the proposed defense, we adopt the Skip-gram model to learn semantic embedding from scratch.

\textbf{Fusion and classification.} 
To improve the model robustness, we leverage intermediate multimodal fusion \cite{williams2018dnn} which is a fine-grained fusion scheme, to incorporate the node representation into semantic representation.
Specifically, we design two unimodal models $\phi^{(g)}(\cdot)$ and $\phi^{(s)}(\cdot)$ for the graph and semantics modalities, respectively.
Hence, for the given adversarial text $\bm{x_{adv}}=\{x'_1, x'_2, \cdots, x'_N\}$, its adversarial representation $E_{g}(\bm{x_{adv}})$ is denoted by the last hidden output of $\phi^{(g)}$, i.e., $E_{g}(\bm{x_{adv}})\!=\!\phi^{(g)}([f(x'_1), f(x'_2), \cdots, f(x'_N)])$, and the semantic representation $E_{s}(\bm{x_{adv}})$ can be obtained similarly.
Then, we concatenate $E_{g}(\bm{x_{adv}})$ and $E_{s}(\bm{x_{adv}})$ to form a semantic-rich representation for the downstream classification tasks, i.e.,
\begin{equation}
    \mathcal{F}(\bm{x_{adv}})=\mathop{\arg\max}_{\hat{y}}\frac{e^{\mathcal{F}_{\hat{y}}(E_{g}(\bm{x_{adv}})\oplus E_{s}(\bm{x_{adv}}))}}{\sum_{i=1}^{C}{e^{\mathcal{F}_{i}(E_{g}(\bm{x_{adv}})\oplus E_{s}(\bm{x_{adv}}))}}},
\end{equation}
where $\mathcal{F}_{i}$ is the confidence of the i-th class, $C$ is the total number of classes, and the defense is viewed as successful if $\hat{y}=y$.

\vspace{-0.15cm}
\section{EXPERIMENT}
\vspace{-0.15cm}
\label{sec:exp}
\subsection{Experimental setting}
\vspace{-0.1cm}
\textbf{Datasets.} 
\system is evaluated on two datasets in the task of classification: (i) Douban Short Movie Comments (DMSC) \cite{dmsc}: It is a public benchmark for Chinese sentiment analysis, consisted of over millions of movie comments. 
(ii) Spam Advertisement (SpamAds): It is a real-world dataset that contains 100,000 user comments collected from the e-commercial platform Taobao, of which 50,000 are spam ads for guiding customers to offline activities that may be risky and the rest are normal comments.
We use DMSC and SpamAds for sentiment analysis and antispam tasks, respectively.

\textbf{Setup.}
We adopt the black-box attack TextBugger \cite{li2019textbugger} to evaluate the efficacy of \system.
We use attack success rate (ASR) and the average number of perturbed words (perturbation) in the adversarial texts to measure the attack performance, and we use semantic similarity \cite{simnet} as well as adversarial similarity (i.e., the maximum of the phonetic and glyph similarity between the original and generated texts) to quantify the quality of generated adversarial texts.
We experiment with two commonly used models, i.e., TextCNN \cite{kim2014convolutional} and BiLSTM \cite{zhou2016attention}, to evaluate the generalizability of \system across architectures.
In addition, we compare \system with a spelling correction-based approach (SC) \cite{pycorrector}.

\begin{table}[t]
    \setlength{\abovecaptionskip}{3pt}
    \caption{Model performance in the non-adversarial scenario. Avg-conf is the average confidence on correctly classified texts.}
    \label{tab:benign_performance}
    \centering
    \scalebox{0.825}{
    \begin{tabular}{l|cc|cc}
         \toprule[1.25pt]
         \multirow{2}*{Model} &  \multicolumn{2}{c|}{Antispam}  & \multicolumn{2}{c}{Sentiment Analysis}\\\cline{2-5}
         & Accuracy & Avg-conf & Accuracy & Avg-conf\\
         \hline
         \hline
         TextCNN          & 0.928 & 0.944 & 0.874 & 0.873\\
         TextCNN+SC       & 0.920 & 0.936 & 0.864 & 0.867\\
         TextCNN+AdvGraph & \bf{0.928} & \bf{0.962} & 0.872 & \bf{0.898}\\
         \hline
         BiLSTM           & 0.893 & 0.894 & 0.851 & 0.849\\
         BiLSTM+SC        & 0.886 & 0.887 & 0.845 & 0.844\\
         BiLSTM+AdvGraph  & \bf{0.914} & \bf{0.937} & \bf{0.864} & \bf{0.847}\\
         \bottomrule[1.25pt]
    \end{tabular}
    }
    \vspace{-0.5cm}
\end{table}

\vspace{-0.25cm}
\subsection{Performance in non-adversarial scenario}
\vspace{-0.1cm}
We first evaluate the effectiveness of \system in the non-adversarial scenario to verify whether the added defense will have a negative impact on the model performance on legitimate texts.

The main results are summarized in \cref{tab:benign_performance}, from which we can see that all the TextCNN and BiLSTM models have achieved considerable high accuracy in the two tasks. 
However, the accuracy is decreased by about 1\% when adopting SC as the defense method.
This is mainly because that the SC method makes erroneous correction over some legitimate texts, resulting in a negative impact on the downstream classification.
In comparison, the accuracy of the models defended by \system is promoted in most cases, which indicates that \system would not affect the model performance in the non-adversarial scenario.
In addition, the average confidence of the models protected by \system on correctly classified samples is higher than that of the common models and the models defended by SC.
It shows that the models defended by \system have learned better decision boundaries which can separate samples of different classes more perfectly.
We argue that this mainly benefits from the semantic-rich representation obtained by multimodal fusion, which introduces more knowledge for the final decision that can hardly be learned by simply increasing the data amount.

\begin{table}[!htb]
    \setlength{\abovecaptionskip}{3pt}
    \caption{Model performance on user-generated obfuscated texts.}
    \label{tab:adversarial_performance}
    \centering
    \scalebox{0.78}{
    \begin{tabular}{l|cc|cc}
         \toprule[1.25pt]
         \multirow{2}*{Model} &  \multicolumn{2}{c|}{Antispam}  & \multicolumn{2}{c}{Sentiment Analysis}\\\cline{2-5}
         & Accuracy & Perturbation & Accuracy & Perturbation \\
         \hline
         \hline
         TextCNN          & 0.630 & 1.23 & 0.669 & 1.16\\
         TextCNN+SC       & 0.758 & 1.47 & 0.734 & 1.25 \\
         TextCNN+AdvGraph & \bf{0.916} & \bf{1.84} & \bf{0.857} & \bf{1.52}\\
         \hline
         BiLSTM           & 0.618 & 1.19 & 0.622 & 1.14\\
         BiLSTM+SC        & 0.743 & 1.41 & 0.715 & 1.22\\
         BiLSTM+AdvGraph  & \bf{0.898} & \bf{1.79} & \bf{0.839} & \bf{1.49}\\
         \bottomrule[1.25pt]
    \end{tabular}
    }
    \vspace{-0.6cm}
\end{table}

\begin{table*}[t]
    \setlength{\abovecaptionskip}{2pt}
    \caption{The attack performance against all the target models under the adaptive setting.}
    \label{tab:adaptive_performance}
    \centering
    \begin{tabular}{l|cccc|cccc}
         \toprule[1.25pt]
         \multirow{2}*{Model} &  \multicolumn{4}{c|}{Antispam}  & \multicolumn{4}{c}{Sentiment Analysis}\\\cline{2-9}
         & ASR & Perturbation & \tabincell{c}{Adversarial\\Similarity} & \tabincell{c}{Semantic\\Similarity} & ASR & Perturbation & \tabincell{c}{Adversarial\\Similarity} & \tabincell{c}{Semantic\\Similarity}\\
         \hline
         \hline
         TextCNN          & 0.769 & 1.63 & 0.917 & 0.874 & 0.703 & 2.07 & 0.911 & 0.832\\
         TextCNN+SC       & 0.763 & 1.56 & 0.919 & 0.873 & 0.673 & 2.02 & 0.902 & 0.831 \\
         TextCNN+AdvGraph & \bf{0.421} & \bf{1.99} & \bf{0.892} & \bf{0.852} & \bf{0.430} & \bf{2.37} & \bf{0.864} & \bf{0.825} \\
         \hline
         BiLSTM           & 0.757 & 1.97 & 0.903 & 0.858 & 0.759 & 2.04 & 0.916 & 0.831\\
         BiLSTM+SC        & 0.738 & 1.92 & 0.931 & 0.872 & 0.716 & 1.99 & 0.910 & 0.837\\
         BiLSTM+AdvGraph  & \bf{0.392} & \bf{2.00} & \bf{0.872} & \bf{0.843} & \bf{0.403} & \bf{2.10} & \bf{0.855} & \bf{0.814}\\
         \bottomrule[1.25pt]
    \end{tabular}
    \vspace{-0.45cm}
\end{table*}

\subsection{Performance in real-world adversarial scenarios}
\vspace{-0.15cm}
We then evaluate the efficacy of \system from the perspective of mitigating the user-generated obfuscated texts.
Specifically, we first collect 1,000 obfuscated spam ads and 1,000 obfuscated negative movie comments from e-commercial platforms and online social media, respectively.
Each collected text is manually confirmed to have at least one variant in the text (as shown in \cref{fig:real_example}).
Then, the efficacy is evaluated on these texts in terms of accuracy and the average number of perturbations in the correctly handled texts.

The evaluation results is shown in \cref{tab:adversarial_performance}. 
It is obviously observed that both the SC-based defense and \system have promoted the model performance on the user-generated obfuscated texts. For instance, the accuracy of TextCNN in the antispam task is promoted by $12.8\%$ when leveraging the SC-based defense and it is promoted by $28.6\%$ when adopting \system.
This indicates that both these two methods are effective in the static adversarial scenario and \system still outperforms the baseline.

\vspace{-0.25cm}
\subsection{Robustness against adaptive adversarial attack}
\vspace{-0.05cm}
Finally, we evaluate the efficacy of \system under the adaptive attack setting assuming attackers know our defense \cite{carlini2017adversarial}.
Under this setting, attackers can explore the vulnerability of the target model and the defense through query access to the whole pipeline.
This is a more realistic worst-case setting since attackers in real adversarial scenarios usually only have black-box query access to the target models, and they would adopt new variation strategies to evade the defense once they perceive the defense.
In this evaluation, we apply TextBugger to mimic the real-world adversaries, and the adversarial texts are generated from 1,000 correctly classified samples randomly sampled for each task.
The maximum perturbation allowed per text is 4 since the average length of sampled texts is about 40.

\begin{figure}[t]
\setlength{\abovecaptionskip}{3pt}
\begin{minipage}[b]{.475\linewidth}
  \centering
  \centerline{\includegraphics[width=1\linewidth]{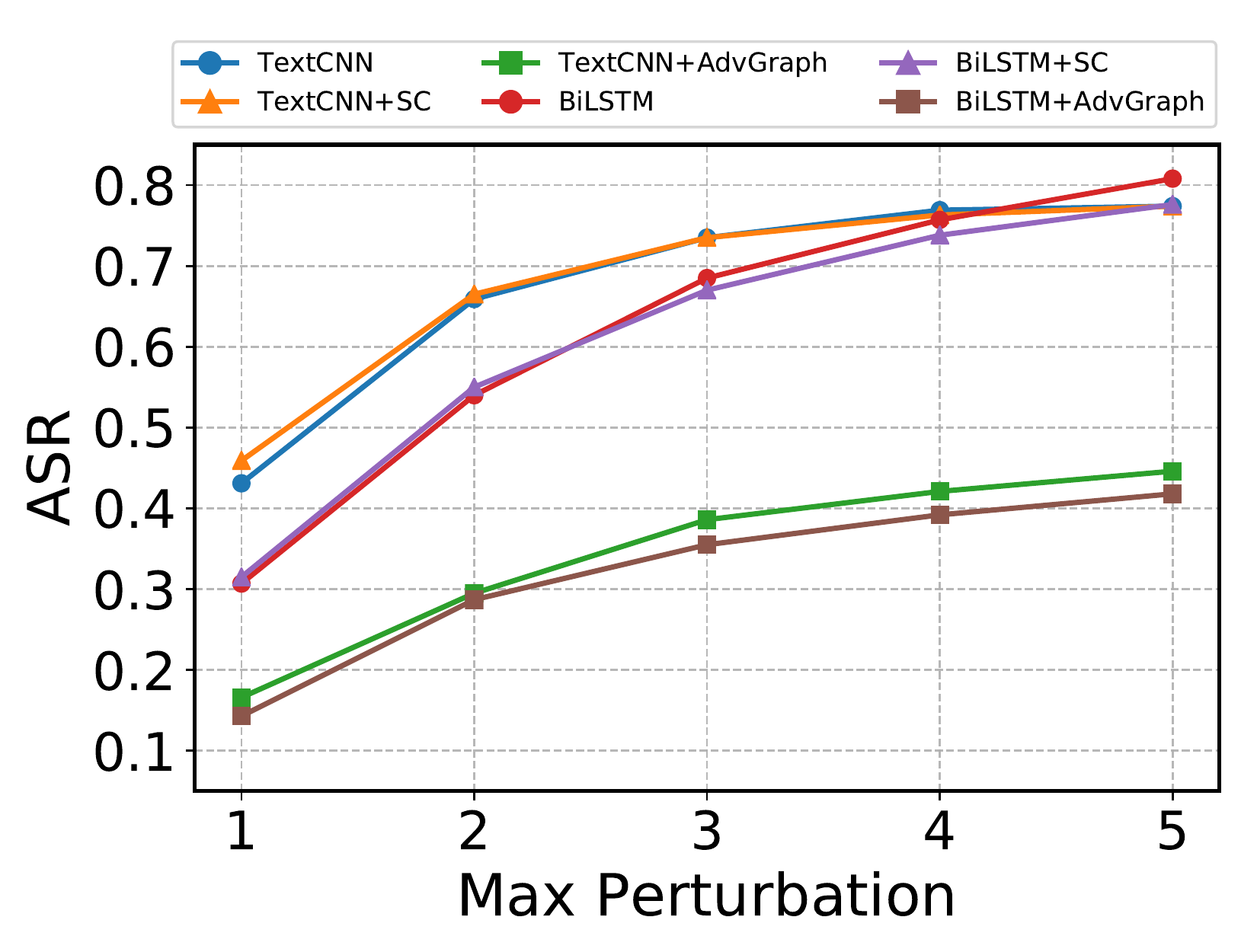}}
  \subcaption{Antispam}
  \label{fig:impact_wx}
\end{minipage}
\hfill
\begin{minipage}[b]{0.475\linewidth}
  \centering
  \hspace*{-0.5cm}
  \centerline{\includegraphics[width=1\linewidth]{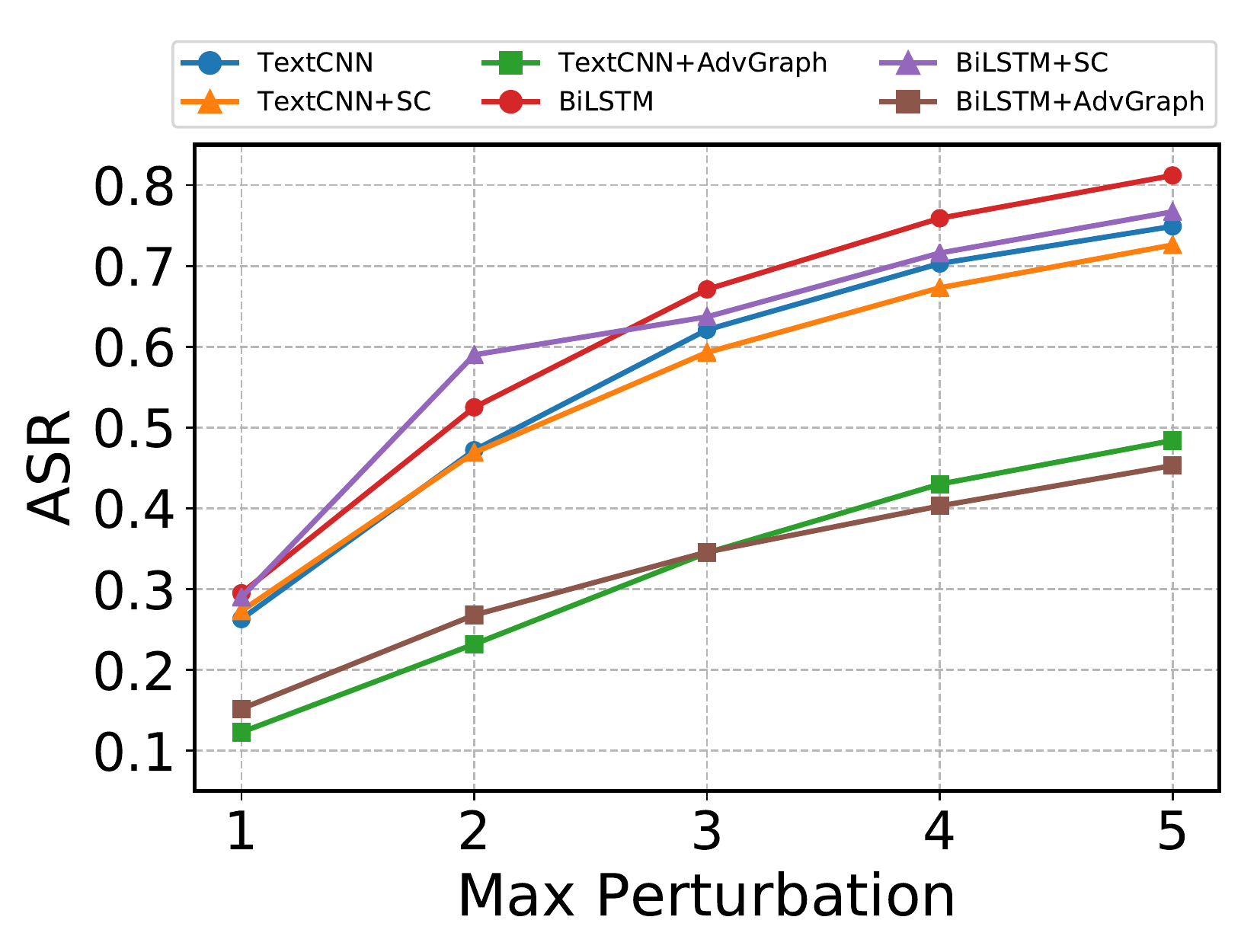}}
  \subcaption{Sentiment Analysis}
  \label{fig:impact_dmsc}
\end{minipage}
\caption{The impact of maximum perturbation allowed on ASR.}
\label{fig:max_perturbation_impact}
\vspace{-0.55cm}
\end{figure}

\textbf{Attack performance.}
The results of adaptive attack are presented in \cref{tab:adaptive_performance}.
Obviously, it is observed that there is only an unnoticeable reduction in ASR against the target models when taking the SC-based method as the defense.
By analyzing the corrected texts output by the SC model, we find that the SC model itself is vulnerable to adaptive attacks while the efficacy of SC-based defense is completely dependent on its correction performance, thus leading to limited effectiveness in the adaptive attack scenario.
On the contrary, the ASR against the models protected by \system is greatly decreased with the required perturbations increased across all the cases, which outperforms the baselines by a significant margin.
In addition, it can be reflected from the adversarial similarity and semantic similarity metrics that the quality of the generated adversarial texts is also worse.
This demonstrates that our proposed defense method is more robust against the adaptive attack and is more effective in weakening the attack threat as well as increasing the attack cost.

\textbf{Impact of maximum perturbation on ASR.}
We also investigate the impact of the maximum perturbations allowed in per text on the ASR against all the target models. 
The analysis results are shown in \cref{fig:max_perturbation_impact}, from which we can see that \system exhibits good performance in mitigating the attack power, i.e., the ASR against the target models defended by \system increases slightly as the allowed maximum number of perturbations grows, and outperforms the baselines by a significant margin. In contrast, the defense efficacy of the SC-based method is negligible, which once again proves that it is not practical in the adaptive adversarial scenario.

\begin{figure}[t]
\setlength{\abovecaptionskip}{3pt}
\begin{minipage}[b]{.475\linewidth}
  \centering
  \centerline{\includegraphics[width=1\linewidth]{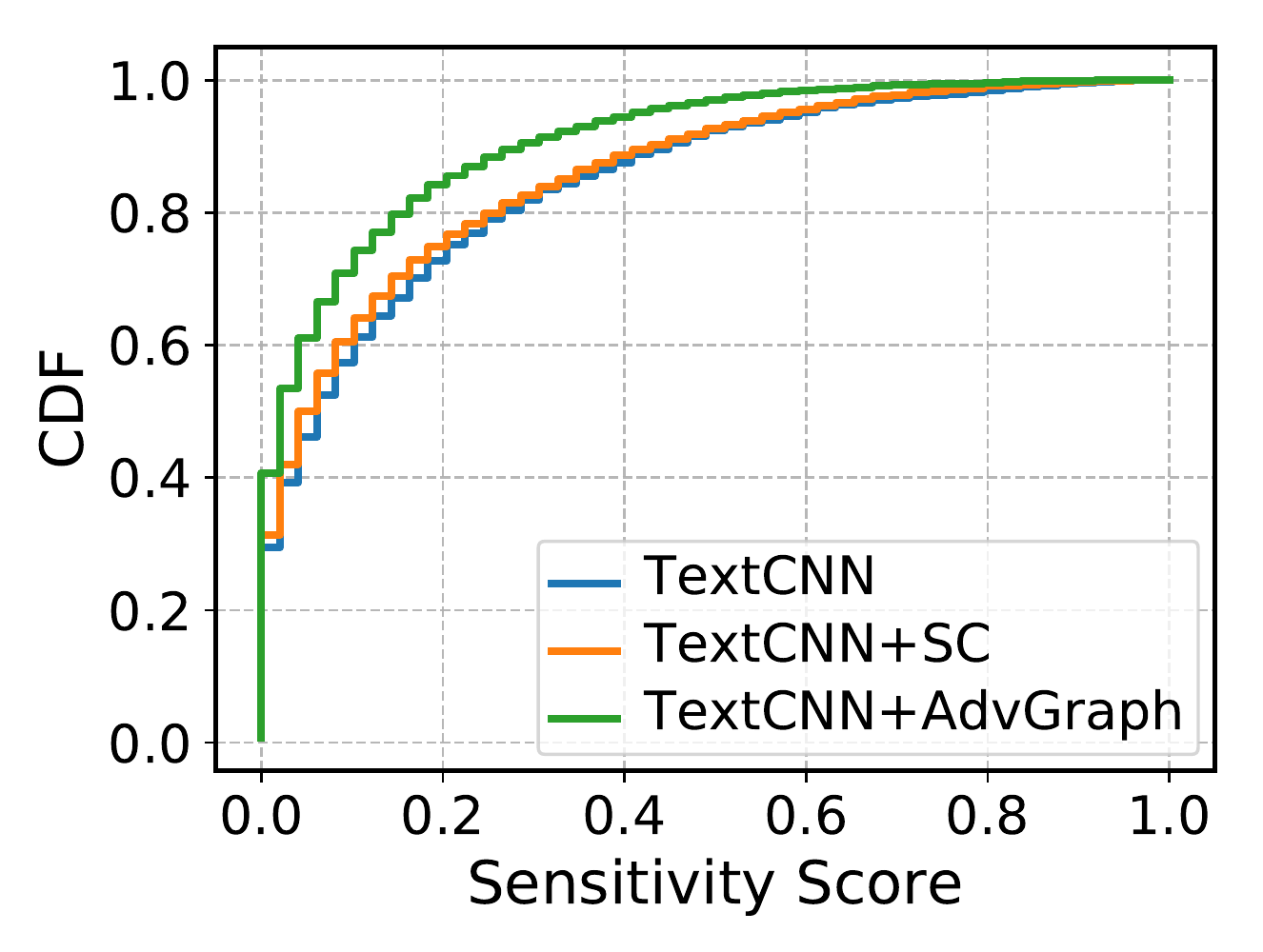}}
  \subcaption{TextCNN}
  \label{fig:textcnn_sensitive}
\end{minipage}
\hfill
\begin{minipage}[b]{0.475\linewidth}
  \centering
  \hspace*{-0.5cm}
  \centerline{\includegraphics[width=1\linewidth]{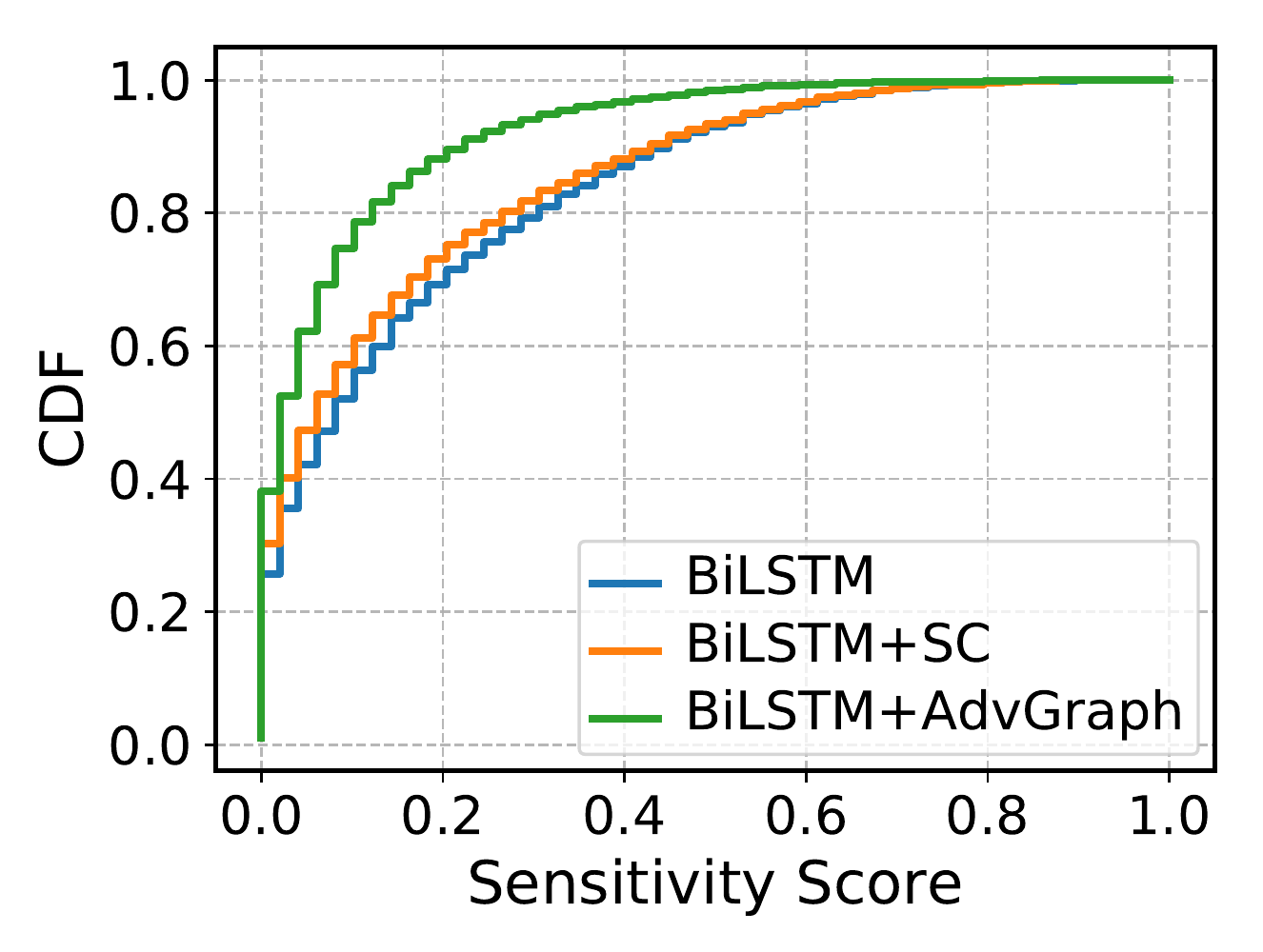}}
  \subcaption{BiLSTM}
  \label{fig:bilstm_sensitive}
\end{minipage}
\caption{The model sensitivity against perturbations in antispam task.}
\label{fig:model_sensitivity}
\vspace{-0.55cm}
\end{figure}

\textbf{Sensitivity analysis.} 
We further analyze the model sensitivity against each perturbation replacement when generating adversarial texts by taking the antispam task as an example. The cumulative distribution of the sensitive score, i.e., the reduction in classification confidence after each perturbation, is visualized in \cref{fig:model_sensitivity}.  
It is clearly observed from \cref{fig:model_sensitivity} that the sensitivity scores of the models defended by \system are much smaller than those of the common models and the models defended by SC, and a more obvious trend is observed on the BiLSTM models.
This indicates that \system does enhance the inherent robustness of the target models, which effectively mitigates their sensitivity to adversarial perturbations.

\vspace{-0.15cm}
\section{CONCLUSION}
\label{sec:conclusion}
\vspace{-0.15cm}
In this paper, we introduce a novel defense specifically designed for the Chinese-based NLP models, which greatly enhances the model robustness by incorporating the adversarial knowledge into the semantic representation of the input with a delicately built adversarial relationship graph.  
The extensive evaluations on two real-world tasks show that \system exhibits excellent performance in defending against the user-generated obfuscated text as well as the adaptive adversarial attacks without negative impact on the model performance over benign inputs.
Although the proposed method is only evaluated on Chinese tasks, we argue that its basic idea can be extended to some other languages like English, and evaluating its generalizability across languages could be a promising future work.

\vfill\pagebreak


\bibliographystyle{IEEEbib}
\bibliography{refs}

\end{document}